%% file: main.tex
\documentclass[11pt]{article}

\usepackage[preprint]{acl}

\usepackage{times}
\usepackage{latexsym}
\usepackage[T1]{fontenc}
\usepackage[utf8]{inputenc}
\usepackage{microtype}
\usepackage{inconsolata}
\usepackage{graphicx}
\usepackage{booktabs}
\usepackage{amsmath}
\usepackage{amssymb}
\usepackage{xcolor}
\usepackage{multirow}
\usepackage{xfp}
\usepackage{float}
\usepackage{capt-of}
\usepackage{algorithm}
\usepackage[noEnd=true, indLines=true, commentColor=black!55]{algpseudocodex}
\algrenewcommand\algorithmicrequire{\textbf{Input:}}
\algrenewcommand\alglinenumber[1]{{\scriptsize\color{black!45}#1}}
\usepackage{tikz}
\usetikzlibrary{positioning,fit,arrows.meta,shapes.geometric}
\usepackage[capitalise,noabbrev,nameinlink]{cleveref}
\crefname{figure}{\mbox{Figure}}{\mbox{Figures}}
\Crefname{figure}{\mbox{Figure}}{\mbox{Figures}}
\crefname{table}{\mbox{Table}}{\mbox{Tables}}
\Crefname{table}{\mbox{Table}}{\mbox{Tables}}
\crefname{algorithm}{\mbox{Algorithm}}{\mbox{Algorithms}}
\Crefname{algorithm}{\mbox{Algorithm}}{\mbox{Algorithms}}
\crefname{appendix}{\mbox{Appendix}}{\mbox{Appendices}}
\Crefname{appendix}{\mbox{Appendix}}{\mbox{Appendices}}
\crefformat{section}{#2\S#1#3}
\Crefformat{section}{#2\S#1#3}
\crefformat{subsection}{#2\S#1#3}
\Crefformat{subsection}{#2\S#1#3}

\newcommand{\var}[1]{\texttt{#1}}
\newcommand{\method}{\mbox{\textbf{RV-ICL}}}
\title{Recursive Video In-Context Learning for Agentic Robot}

\author{Wenrui Bao\textsuperscript{1} \quad Xinxin Liu\textsuperscript{1} \quad Bingxin Xu\textsuperscript{2} \quad Yuzhang Shang\textsuperscript{1}\thanks{Corresponding author.} \\
  \textsuperscript{1}University of Central Florida \quad \textsuperscript{2}University of Southern California \\
  \texttt{wenrui.bao@ucf.edu} \\
  {\small \href{https://bwr-hhh.github.io/rvicl/}{Project page} \quad \href{https://github.com/BWR-hhh/rvicl}{Code}}}

\begin{document}
\maketitle

\input{sections/abstract}
\input{sections/introduction}
\input{sections/related_work}
\input{sections/method}
\input{tables/main_results}
\input{sections/experimental_setup}
\input{sections/results}
\input{sections/conclusion}
\input{sections/limitations}
\input{sections/ethics}

\bibliography{custom}

\appendix
\input{sections/appendix}

\end{document}

%% file: sections/abstract.tex
\begin{abstract}
LLM agents that orchestrate frozen vision-language-action (VLA) policies improve across episodes through \emph{text} memory, which records what the agent did but not how the task is done. A demonstration video shows it, but fits poorly into an agent's context. The full video slows every turn, fixed keyframes lose the contact detail that decides whether a grasp holds, and what the agent needs shifts from the task's structure while planning to the frames around each contact. We introduce \textbf{Recursive Video In-Context Learning (\method{})}, a training-free method that turns a demonstration into a hierarchy the agent navigates rather than a prompt it receives. The hierarchy is built from the sub-events of the demonstration, such as grasps and releases. Its levels grow finer, from keyframes of the whole task to phases, moments and short clips, and are exposed through read-only tools. The agent reads the coarse levels before planning. During execution it re-enters the hierarchy whenever a step needs more detail and loads only the clip of its current sub-goal. One demonstration per task is enough. Built on RPent, \method{} raises success from 92.6\% to 96.5\% on LIBERO-PRO and from 86.7\% to 95.8\% on LIBERO-Plus.
\end{abstract}


%% file: sections/introduction.tex
\begin{figure*}[t]
\centering
\includegraphics[width=\textwidth]{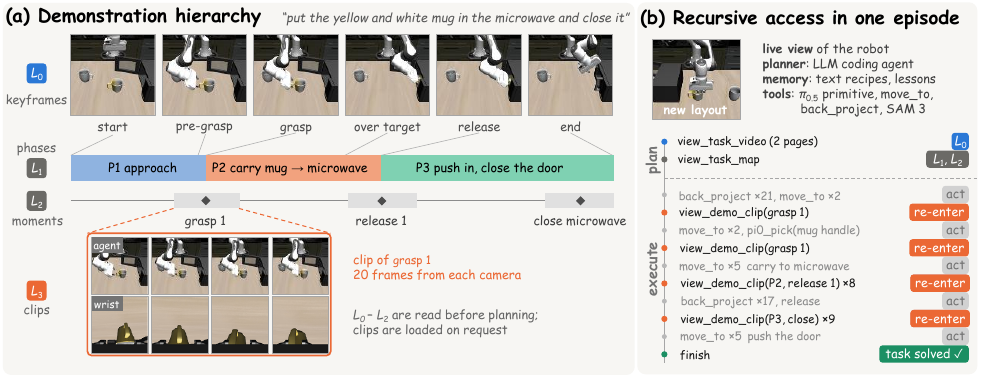}
\caption{Overview of \method{} on a task from the LIBERO-10 suite of LIBERO-PRO. The agent receives one demonstration of the task as a hierarchy (a) and consults it while solving the same task in a new layout (b). (a) The hierarchy has four levels: $L_0$ event-driven keyframes (thin lines mark their time), $L_1$ phases between grasps and releases, $L_2$ moments around each sub-event, and $L_3$ 20-frame clips, here of the first grasp. (b) The tool calls of one solved evaluation episode (consecutive calls grouped, robot actions in grey, re-entries in orange). The agent reads $L_0$--$L_2$ while planning and loads clips only during execution. Each re-entry loads the clip of the current sub-goal: the grasp, the release or the closing of the door.}
\label{fig:tree}
\vspace{-0.2in}
\end{figure*}

\section{Introduction}

Large language models (LLMs) are increasingly deployed as the deliberative layer of robot systems. They decompose instructions, localise objects, and call learned visuomotor policies for the contact-rich steps they cannot perform analytically~\citep{ahn2022saycan,liang2023cap,huang2022innermonologue}. Recent agentic harnesses push this design further. HarnessVLA~\citep{zhang2026harnessvla}, released as the RPent framework, exposes a frozen VLA as one retryable primitive among a few analytic ones. A coding agent orchestrates them and improves across episodes through \emph{text memory}: recipes, lessons and failure rules. Under distribution shift, this design outperforms end-to-end VLAs by a wide margin.

How far the agent can improve, however, is bounded by what text can record. The agent writes its memory from its own execution traces, so the memory describes its calls and their outcomes rather than the task itself. It can record that a grasp failed, but not what a successful execution looks like: which object is taken first, from which side the hand approaches, how high the object is lifted, when the gripper opens. Such knowledge is spatial and temporal, and it is hard to put into words.

What is hard to put into words is easy to show: a video of the task captures all of it, frame by frame. Such a video is also cheap to obtain. A human demonstration only requires recording a person actually performing the task, with no action labels or other annotation. Large models can learn from demonstrations placed in their context, in text~\citep{brown2020gpt3} and with interleaved images~\citep{alayrac2022flamingo}. This suggests letting the agent watch a demonstration of its task, much as a person watches a how-to video before trying something new.

But in what form should the video enter the agent's memory? The obvious answers fail for three reasons. \textbf{(i) Length.} A demonstration spans hundreds of frames, and the agent re-reads its whole context at each of its tens to hundreds of turns. Long contexts also degrade how well models use their content~\citep{liu2024lost}. \textbf{(ii) Resolution.} A few keyframes are affordable, but they drop the frames around each contact, which decide whether a grasp holds or a placement registers. \textbf{(iii) Timing.} What the agent needs changes over the episode: the task's structure before it acts, and the frames around a particular contact once it reaches that step. No fixed presentation of the video can satisfy all three constraints at once.

Since each moment of the episode calls for a different part of the demonstration, why not let the agent watch just that part, and only at that moment? Methods for long documents and long videos already take this route: instead of reading everything, the language model \emph{navigates} a hierarchical index~\citep{sarthi2024raptor,wang2024videoagent,packer2023memgpt}. We bring this idea to embodied agents, which decide from their own progress when to look closer: before a delicate contact, at the start of a new sub-goal, or after a step goes wrong.

We build such an index over a demonstration and call the method \textbf{Recursive Video In-Context Learning (\method{})}. The index is a tree whose levels grow finer: event-driven keyframes of the whole task, the phases between sub-events, the moments around each sub-event, and short clips from any camera (\cref{fig:tree}a). The agent reads the coarse levels before planning and does not collect clips up front. Whenever it needs more detail during execution, the agent \emph{re-enters} the tree at the node of its current sub-goal and pulls only the relevant clip (\cref{fig:tree}b). The demonstration is thus consulted recursively, at the level of detail and the moment the agent's progress calls for.

\method{} requires no training, no change to the VLA, and one demonstration per base task. The hierarchy is built from the sub-events of the demonstration, the instants at which one sub-task ends and the next begins. The same construction therefore applies to any task that decomposes into sub-tasks. We make the following contributions:
\begin{itemize}
\itemsep0em
\item We introduce \emph{video in-context learning} to tool-using robot agents: their text memory cannot record what a successful execution looks like, but a demonstration video shows it frame by frame (\cref{sec:setting}).
\item We propose Recursive Video In-Context Learning (\method{}), which lets the agent look only at the part of the demonstration its current step needs, by giving it \emph{recursive access} to a hierarchy built from the demonstration's sub-events (\cref{sec:method}).
\item On LIBERO-PRO, \method{} raises the success rate of HarnessVLA from 92.6\% to 96.5\%, above every baseline, and on LIBERO-Plus from 86.7\% to 95.8\% across seven kinds of perturbation. On the long LIBERO-10 tasks it gains up to 12 points, where keyframes or the whole video of the same demonstration gain at most 2 (\cref{sec:results}).
\end{itemize}

%% file: sections/related_work.tex
\section{Related Work}
\label{sec:related}

\subsection{LLM agents for robot manipulation}
Language models entered robotics as planners over pre-trained skills. SayCan~\citep{ahn2022saycan} scores LLM proposals by learned affordances, ProgPrompt~\citep{singh2023progprompt} and Code as Policies~\citep{liang2023cap} write programs over perception and control APIs, and VoxPoser~\citep{huang2023voxposer} composes 3D value maps for a motion planner. Inner Monologue~\citep{huang2022innermonologue} feeds execution feedback back to the planner, and Voyager~\citep{wang2023voyager} accumulates a library of verified skills.

A recent line of \emph{agentic harnesses} puts a coding agent in charge of a frozen VLA. HarnessVLA~\citep{zhang2026harnessvla} exposes the VLA as a retryable primitive and learns its operating range from text memory; Robo-Harness K1~\citep{li2026robokharness} augments such agents with perception; CaP-X~\citep{fu2026capx} benchmarks coding agents for manipulation; RATS~\citep{zhang2026rats} learns through play; and RAPID~\citep{liu2026rapid} synthesises agent programs from demonstrations.

The knowledge these systems carry across episodes is textual or programmatic. \method{} keeps the harness unchanged and adds a visual record of how the task is performed, together with a protocol for consulting it.

\subsection{Learning from demonstrations in context}
In language models, a few demonstrations in the prompt suffice to specify a task~\citep{brown2020gpt3}, although models draw mostly on their format and label space rather than the exact input--label mapping~\citep{min2022rethinking}. Flamingo~\citep{alayrac2022flamingo} extends in-context learning to interleaved images.

In robotics, one-shot imitation has been learned by meta-learning from human videos~\citep{yu2018daml} and by video-conditioned policies~\citep{jain2024vid2robot}. In-context imitation has been achieved by tokenising demonstrations as keypoints for an off-the-shelf LLM~\citep{dipalo2024kat}, by training a policy to continue interleaved demonstration trajectories~\citep{fu2024icrt}, and by conditioning a VLA on structured demonstrations~\citep{xu2026stellavla}. When an LLM acts directly as the policy, one-shot demonstrations yield no aggregate gain~\citep{zhang2026astrarobodojo}.

All of these present the demonstration once, as a prefix for the next actions. Our agent makes tens to hundreds of decisions per episode, so the question is not only \emph{whether} to show a demonstration but \emph{which part, and when}.

\subsection{Recursion in language-model reasoning and retrieval}
\label{sec:related-recursion}
Recursion lets a bounded model handle inputs and problems that exceed its context. \emph{Recursive summarisation} condenses a book into a tree of summaries read top-down~\citep{wu2021recursively}, and RAPTOR~\citep{sarthi2024raptor} retrieves from such a tree at the level of abstraction a question requires. For long videos, VideoAgent~\citep{wang2024videoagent} and VideoTree~\citep{wang2025videotree} likewise inspect only the frames or branches a question needs.

\emph{Recursive decomposition} splits a problem as needed. ADaPT~\citep{prasad2024adapt} decomposes a sub-task further only when the executor fails on it, THREAD~\citep{schroeder2025thread} spawns child threads that return only what the parent needs, and Recursive Language Models~\citep{zhang2025rlm} call themselves over slices of a context too long to read at once. Tree of Thoughts~\citep{yao2023tot} searches over such trees of intermediate steps.

A related family decides \emph{when} to fetch information. FLARE~\citep{jiang2023flare} retrieves when upcoming tokens are uncertain, and Self-RAG~\citep{asai2024selfrag} learns to request retrieval on demand. Reflexion~\citep{shinn2023reflexion} and Self-Refine~\citep{madaan2023selfrefine} feed the outcome of an attempt into the next one, and MemGPT~\citep{packer2023memgpt} pages information between a bounded context and external storage.

\method{} brings recursion to an embodied agent. The demonstration is a tree read at the level each decision requires, and the agent descends whenever its progress in the physical task calls for more detail: before a delicate contact, at a new sub-goal, or after a step goes wrong.

%% file: sections/method.tex
\section{Recursive Video In-Context Learning}
\label{sec:method}

\begin{figure*}[t]
\centering
\includegraphics[width=\textwidth]{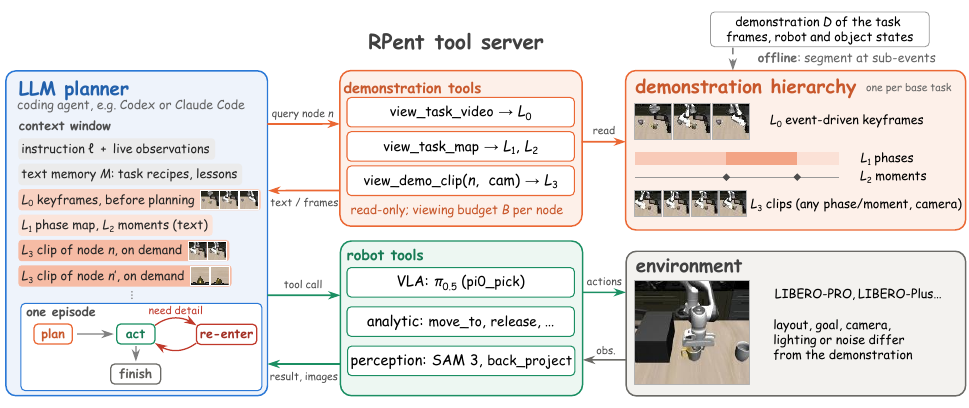}
\vspace{-0.2in}
\caption{Architecture of \method{} on RPent. Each demonstration is segmented offline at its sub-events into a four-level hierarchy. During an episode the LLM agent acts through the harness. Robot tools execute on the environment, and read-only demonstration tools load a level of the hierarchy into the agent's context. The agent reads the coarse levels ($L_0$--$L_2$) before planning and loads a clip whenever its current sub-goal needs more detail.}
\label{fig:arch}
\vspace{-0.2in}
\end{figure*}

\subsection{Problem setting}
\label{sec:setting}
An agent $\pi$, an LLM running as a coding agent, receives an instruction $\ell$ and interacts with a robot environment through a tool set $\mathcal{T}_R$ (\cref{fig:arch}). $\mathcal{T}_R$ contains a VLA primitive for contact-rich steps (\var{pi0\_pick}, which calls $\pi_{0.5}$), analytic primitives for staging, transport and release (\var{move\_to}, \var{release}), perception (SAM~3 segmentation and \var{back\_project}, which lifts an image pixel to a 3D point), and \var{finish}.

At turn $t$ the agent conditions on its context $c_t$ and emits a call $a_t = \pi(c_t)$. The context holds $\ell$, the text memory $M$ of HarnessVLA, the current camera images and the history of calls and results. The episode ends when the agent calls \var{finish} or reaches its step or time limits. It succeeds if the simulator's goal predicate holds.

\method{} adds a demonstration $D$ of the same base task, recorded in a different scene, and a second, read-only tool set $\mathcal{T}_D$ whose calls return views of $D$. The design question is which views $\mathcal{T}_D$ offers and when the agent should request them, given that every view stays in $c_t$ once loaded.

\subsection{The demonstration hierarchy}
\label{sec:store}
A demonstration $D = (o_{1:T}, x_{1:T}, g_{1:T}, s_{1:T})$ consists of multi-camera frames $o_t$, end-effector poses $x_t$, a binary gripper command $g_t$ and, when available, object and joint states $s_t$. We segment $D$ at its \emph{sub-events} $E$, the instants at which one sub-task ends and the next begins. Every task that decomposes into sub-tasks has such instants, so the construction below applies to it unchanged. Manipulation tasks expose their sub-events in the recorded state. An object is grasped when the gripper closes and released when it opens, and a drawer or door is opened or closed when its joint moves. We therefore take $E$ to be the gripper events $E_g = \{t : g_t \neq g_{t-1}\}$ together with the instants at which an articulated fixture opens or closes. The hierarchy has four levels (\cref{fig:tree}a).

\paragraph{$L_0$: task keyframes.} A fixed number $K$ of frames that span the whole task. They are placed around every grasp and release, and uniform samples fill the rest. Each keyframe shows the third-person and wrist views and is captioned with its time fraction, gripper state, end-effector position and event label.

\paragraph{$L_1$: phases.} The intervals $p_j = [e_j, e_{j+1})$ between consecutive gripper events $e_j \in E_g$, rendered as \emph{text}. Each phase records its duration, the gripper state, the end-effector travel and lift, and, when object states are recorded, what it achieved: the object it moved and the region it ended in, or the fixture it opened or closed. A fixture event does not split a phase and is reported as part of what that phase achieved. The task's goal predicates are listed with the phases, so each predicate can be matched to the phase that satisfies it.

\paragraph{$L_2$: moments.} A text index of short windows around every sub-event and at the end of the episode. Each window is named by kind and order (\textit{grasp 1}, \textit{release 1}, \textit{close microwave}).

\paragraph{$L_3$: clips.} A short sequence of frames sampled evenly from any phase or moment, from the third-person view, the wrist view or the two stacked.

The hierarchy conveys \emph{how} the task is done, not \emph{where}. The demonstration was recorded in another scene, so the tool descriptions tell the agent to re-localise every object in its own cameras rather than read positions from the demonstration.

\subsection{Recursive access}
\label{sec:access}
$\mathcal{T}_D$ exposes the hierarchy through three calls. \var{view\_task\_video}$(\textit{page})$ returns $L_0$ one page of keyframes at a time, \var{view\_task\_map}$()$ returns $L_1$ and the $L_2$ index, and \var{view\_demo\_clip}$(n, \textit{cam})$ returns the $L_3$ clip of a phase or moment $n$. The system prompt tells the agent how to use the three calls, and the agent decides when to make them. Access is recursive: the agent reads the coarse levels once and then, at each step that needs detail, returns to the hierarchy for one clip (\cref{alg:rvicl}).

\paragraph{Coarse levels before planning.}
Before its first action the agent reads $L_0$ to $L_2$. It writes its sub-goals from the phase map, with the keyframes as the picture of each stage. It loads no clip yet, because the prompt asks for clips at the point of action rather than up front.

\paragraph{Re-entry during execution.}
The agent returns to the hierarchy whenever its current sub-goal calls for more detail. It does so before it closes the gripper on an object or opens it, and when a stretch of motion is unclear. It also does so when a step does not go as planned, such as a grasp that does not close, an object that does not move, or a placement whose goal predicate does not fire. In that case it watches the clip before it changes any parameter or primitive. It does not reread the coarse levels. It maps the current sub-goal $g$ to its phase $p_j$ and loads the clip of $p_j$ or of the moment it needs. It compares the clip with its current views and adjusts its next action. A re-entry adds one clip to the context, not the whole video.

\paragraph{Bounding the recursion.}
The harness can bound the recursion with a viewing budget $B$ per phase or moment. Further requests for the same phase or moment are refused with a message that sends the agent back to acting. A finite budget stops the agent from re-watching a step that a clip cannot fix, as in the failed episodes of \cref{sec:recurse}.

\begin{algorithm}[!]
\caption{One \method{} episode. $\oplus$ appends to the context, and the budget $B$ caps the views of each phase or moment.}
\label{alg:rvicl}
\small
\begin{algorithmic}[1]
\Require instruction $\ell$, memory $M$, demonstration $D$
\State $c \gets [\ell, M, o_{\mathrm{now}}]$
\State $c \gets c \oplus L_0(D) \oplus L_1(D) \oplus L_2(D)$ \Comment{coarse levels}
\State $\mathit{plan} \gets \pi.\textsc{Plan}(c)$ \Comment{sub-goals from the phases}
\For{sub-goal $g$ in $\mathit{plan}$}
  \While{$g$ not satisfied and steps remain}
    \If{$\pi$ needs more detail on $g$} \Comment{re-enter}
      \State $n \gets \textsc{Node}(g, D)$
      \If{$\textsc{Views}(n) < B$}
        \State $c \gets c \oplus L_3(n)$
      \EndIf
    \EndIf
    \State $a \gets \pi(c)$
    \State $c \gets c \oplus \textsc{Execute}(a)$
  \EndWhile
\EndFor
\State \Return $\pi.\textsc{Finish}(c)$
\end{algorithmic}
\end{algorithm}
\vspace{-0.1in}

%% file: tables/main_results.tex
\begin{table*}[t]
\centering\small
\setlength{\tabcolsep}{4.2pt}
\begin{tabular}{@{}llccccccccc@{}}
\toprule
& Method & Spat-T & Spat-S & Obj-T & Obj-S & Goal-T & Goal-S & L10-T & L10-S & Overall \\
\midrule
\multirow{8}{*}{\rotatebox{90}{\scriptsize VLA}}
& OpenVLA~\citep{kim2024openvla} & 0.0 & 0.0 & 0.0 & 0.0 & 0.0 & 0.0 & 0.0 & 0.0 & 0.0 \\
& $\pi_0$~\citep{black2025pi0} & 0.0 & 0.0 & 0.0 & 2.0 & 0.0 & 0.0 & 0.0 & 0.0 & 0.3 \\
& $\pi_{0.5}$~\citep{black2025pi05} & 1.0 & 20.0 & 1.0 & 17.0 & 2.0 & 38.0 & 1.0 & 8.0 & 11.0 \\
& MolmoAct~\citep{lee2025molmoact} & 0.0 & 0.0 & 0.0 & 6.0 & 0.0 & 0.0 & 6.0 & 0.0 & 1.5 \\
& NORA~\citep{hung2025nora} & 0.0 & 0.0 & 0.0 & 0.0 & 0.0 & 0.0 & 0.0 & 0.0 & 0.0 \\
& X-VLA~\citep{zheng2026xvla} & 0.0 & 0.0 & 8.0 & 2.0 & 9.0 & 1.0 & 10.0 & 0.0 & 3.8 \\
& AtomVLA~\citep{sun2026atomvla} & 1.0 & 16.0 & 0.0 & 10.0 & 11.0 & 2.0 & 9.0 & 1.0 & 6.3 \\
& $\pi_{\mathrm{RLinf}}$~\citep{yu2025rlinf} & 42.0 & 59.0 & 71.0 & 78.0 & 45.0 & 42.0 & 49.0 & 14.0 & 50.0 \\
\midrule
\multirow{5}{*}{\rotatebox{90}{\scriptsize Agent}}
& Cap-X~\citep{fu2026capx} & 14.0 & 12.0 & 18.0 & 22.0 & 17.0 & 26.0 & / & / & 18.2 \\
& RATS~\citep{zhang2026rats} & 31.0 & 29.0 & 63.0 & 61.0 & 36.0 & 43.0 & / & / & 43.8 \\
& HarnessVLA (Codex, GPT-5.5) & 81.0 & 69.0 & 94.0 & 91.0 & 75.0 & 66.0 & 52.0 & 49.0 & 72.1 \\
& HarnessVLA (Claude Code, Opus-4.7) & 94.0 & 80.0 & 88.0 & 90.0 & 87.0 & 87.0 & 71.0 & 62.0 & 82.4 \\
& HarnessVLA (Codex, GPT-6 Astra, low) & \textbf{100.0} & 98.0 & \textbf{100.0} & 99.0 & 88.0 & \textbf{99.0} & 85.0 & 72.0 & 92.6 \\
\midrule
& \textbf{+ \method{}} (Codex, GPT-6 Astra, low) & \textbf{100.0} & \textbf{100.0} & \textbf{100.0} & \textbf{100.0} & \textbf{100.0} & 98.0 & \textbf{90.0} & \textbf{84.0} & \textbf{96.5} \\
\bottomrule
\end{tabular}
\caption{Success rate (\%) on LIBERO-PRO under instruction redirection (T) and position swap (S). Baseline cells aggregate 100 episodes (10 tasks $\times$ 10 seeds) and our cells 50 episodes (10 tasks $\times$ 5 seeds). All rows except ours are reported by \citet{zhang2026harnessvla}; ``/'' marks cells the original works do not report, and the overall score of Cap-X and RATS averages their six reported cells. HarnessVLA uses $\pi_{0.5}$ fine-tuned by RLinf as its VLA primitive; \method{} adds the hierarchy to its GPT-6 Astra configuration. Bold marks the best result in each column.}
\label{tab:main}
\vspace{-0.2in}

\end{table*}

%% file: sections/experimental_setup.tex
\section{Experimental Setup}
\label{sec:setup}

We first test whether \method{} improves a strong agentic system under distribution shift (\cref{sec:main}) and whether recursive access beats showing the agent keyframes or the whole video (\cref{sec:ablation}). We then examine how agents use the hierarchy (\cref{sec:recurse}) and whether the gains hold on a second benchmark with other kinds of perturbation (\cref{sec:plus}).

\paragraph{Benchmarks.}
Our main benchmark is \textbf{LIBERO-PRO}~\citep{zhou2025liberopro}, which perturbs the four LIBERO suites (Spatial, Object, Goal, LIBERO-10; \citealp{liu2023libero}) by instruction redirection (T) and position swap (S). Each of the eight cells contains 10 tasks. We follow the evaluation protocol of \citet{zhang2026harnessvla}, whose reported baselines use 10 seeds per task. We evaluate \method{} with 5 seeds per task, so each of our cells aggregates 50 episodes. To test other kinds of perturbation we use \textbf{LIBERO-Plus}~\citep{fei2025liberoplus}, which perturbs the LIBERO tasks along seven dimensions: object layout, camera viewpoint, robot initial state, language, lighting, background and sensor noise. LIBERO-Plus contains 10,030 perturbed variants, so we evaluate a stratified subset of 120. It takes three variants of each of the 40 base tasks and rotates the seven dimensions over them, which gives 17 or 18 variants per dimension. Variants are drawn from the hardest difficulty levels available, and 105 of the 120 come from the two hardest of the five levels. Both methods use the same initial state for every variant. The agent receives the text memory of the unperturbed base task. On LIBERO-Plus we run HarnessVLA ourselves with the same agent. Since this protocol differs from the official one, these numbers compare the two methods with each other, not with the public leaderboard.

\paragraph{Demonstrations.}
We build one hierarchy per LIBERO base task from the first of its human teleoperation demonstrations. The keyframes are re-rendered at 512\,px from the recorded simulator states, and the clips use the 128\,px frames recorded during teleoperation. Both are shown in the orientation of the agent's own cameras (third-person and wrist), so that demonstration and live views are directly comparable. Replaying the recorded states also gives the object and joint states that $L_1$ uses to state what each phase achieved. Both benchmarks use $K{=}16$ keyframes, and a clip holds at most 20 frames (\cref{app:hier}). One hierarchy serves every perturbed variant of its base task in LIBERO-PRO and LIBERO-Plus. In the instruction-redirection (T) cells, whose goal differs from the demonstrated one, the agent is told which parts of the demonstration no longer apply.

\paragraph{Baselines.}
On LIBERO-PRO we compare with eight VLAs (OpenVLA, $\pi_0$, $\pi_{0.5}$, MolmoAct, NORA, X-VLA, AtomVLA and $\pi_{0.5}$ fine-tuned with RLinf), two agentic systems (Cap-X and RATS), and HarnessVLA with three planners. All baseline numbers are taken from \citet{zhang2026harnessvla}. We define the ablation arms in \cref{sec:ablation}.

\paragraph{Metrics.}
The primary metric is task success rate, read from the simulator's own goal check. On LIBERO-PRO the baseline numbers come from separate runs, so we test the gap in success rate pooled over the eight cells with a two-proportion z-test (\cref{sec:main}). To study how agents use the hierarchy, we log every demonstration call with its level, the phase or moment it shows, its camera and its turn. We also measure the cost of an episode in wall-clock time and tokens (\cref{app:cost}).

\paragraph{Implementation.}
The agent is \var{gpt-6-astra} at low reasoning effort running in the Codex CLI, as in the strongest HarnessVLA configuration. \method{} does not depend on this choice. It only adds read-only tools and a prompt section to the harness and changes nothing specific to the model, so any agent the harness supports can use the same demonstration hierarchy, for example Claude Code with Claude Opus 5.5 (\cref{app:planner}). The VLA primitive is $\pi_{0.5}$ fine-tuned by RLinf on LIBERO and served locally, and SAM~3~\citep{carion2026sam3} provides segmentation. Each episode is a single attempt.

%% file: sections/results.tex
\section{Results}
\label{sec:results}

\subsection{Main results}
\label{sec:main}
\begin{figure*}[t]
\centering
\includegraphics[width=\textwidth]{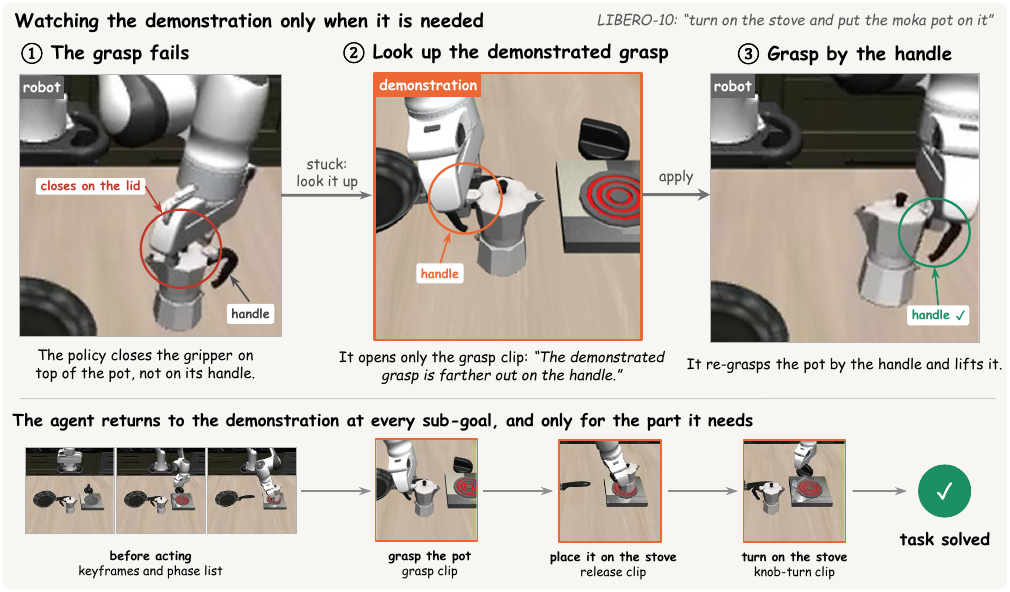}
\caption{How the agent uses the demonstration in one LIBERO-10 episode from our LIBERO-PRO runs. Top: the policy closes the gripper on top of the moka pot (1). The agent does not replay the whole video. It opens only the grasp clip, where the demonstrator takes the pot by its handle (2), and then re-grasps the pot by the handle (3). Bottom: across the episode the agent reads the keyframes and the phase map before acting, then opens one short clip at each sub-goal. Robot views are crops of the episode video, demonstration views are frames of the demonstration, and the quote is the agent's own message.}
\label{fig:recursion}
\vspace{-0.2in}

\end{figure*}

\Cref{tab:main} reports success on the eight \mbox{LIBERO-PRO} cells for end-to-end VLAs, agentic systems and \method{}. HarnessVLA lets a GPT-6 Astra agent call the end-to-end VLA $\pi_{\mathrm{RLinf}}$ as a tool and raises its success by 42.6 points to 92.6\%. \method{} adds recursive access to demonstration clips on top of HarnessVLA and raises success by another 3.9 points to 96.5\%. A two-proportion \mbox{z-test} on the pooled cells shows that this gain is significant (\mbox{$z=2.97$}, \mbox{two-sided $p=0.003$}).

While keeping the other five cells at or near the ceiling, \method{} brings clear gains on the three cells where HarnessVLA is weakest. Goal-T rises from 88.0 to 100.0, L10-T from 85.0 to 90.0 and L10-S from 72.0 to 84.0. LIBERO-10 is the long-horizon suite of the benchmark, and each of its tasks chains several grasps and releases. The gains there suggest that recursive access is especially effective on such tasks. The small drop on \mbox{Goal-S} (98.0 vs.\ 99.0) comes from a single failed episode. With five seeds per task, one episode moves a cell by two points. So the drop does not indicate that \method{} is worse than HarnessVLA on \mbox{Goal-S}.

\subsection{Ablations}
\label{sec:ablation}

We ablate the form in which the demonstration enters the agent's context. All arms share the agent and the text memory and run on the two \mbox{LIBERO-10} cells, which hold the longest tasks:
\begin{itemize}
\itemsep0em
\item \textbf{Text}: HarnessVLA with text memory only.
\item \textbf{Keyframes}: only the event-driven keyframes of the demonstration ($L_0$).
\item \textbf{Full video}: the whole demonstration video, with no hierarchy.
\item \method{}: the full hierarchy ($L_0$--$L_3$) with recursive access (\cref{sec:access}).
\end{itemize}
\Cref{tab:ablation} reports the results. The Text and \method{} rows are taken from the LIBERO-10 cells of \cref{tab:main}. Showing the demonstration is not enough on its own. Keyframes raise success over text memory by only 1.0 point on L10-T and 2.0 points on L10-S. The full video is 1.0 point below text memory on L10-T and 6.0 points below it on L10-S. \method{} is the best arm on both cells, with 90.0\% on L10-T and 84.0\% on L10-S. Its lead is largest on L10-S, the harder cell for every arm, where it is 10.0 points above keyframes and 18.0 points above the full video.

\vspace{-0.0094in}

All three video arms show the same demonstration in different forms. Keyframes are coarse. They show where the gripper closes and opens but not how a grasp or a release is carried out. They are also what \method{} reads before planning, so the gap between the two arms comes from the phase map and the clips. The full video holds that detail but has no index into it, so the agent must find the moment it needs among unrelated frames. \method{} keeps the detail and adds the index, whose text levels tell the agent which clip to ask for. The agent reads the coarse levels first and opens a clip only when a sub-goal calls for it, so the detail it sees is the detail it is about to use. The gap to the full video suggests that what helps is not seeing more but seeing the right part at the right time.

\begin{table}[t]
\centering\small
\setlength{\tabcolsep}{3pt}
\begin{tabular}{@{}llccc@{}}
\toprule
Arm & Demonstration & Re-entry & L10-T & L10-S \\
\midrule
Text & none & -- & 85.0 & 72.0 \\
Keyframes & $L_0$ keyframes & -- & 86.0 & 74.0 \\
Full video & whole video & -- & 84.0 & 66.0 \\
\method{} & $L_0$--$L_3$ & \checkmark & \textbf{90.0} & \textbf{84.0} \\
\bottomrule
\end{tabular}
\caption{Ablations on the two LIBERO-10 cells of LIBERO-PRO, the longest-horizon tasks (GPT-6 Astra, success \%). $L_0$ denotes the keyframes and $L_3$ the clips of the demonstration hierarchy (\cref{sec:store}). Bold marks the best result in each column.}
\label{tab:ablation}
\vspace{-0.25in}
\end{table}

\subsection{How agents recurse}
\label{sec:recurse}

\Cref{fig:recursion} shows what a re-entry gives the agent. After the failed grasp it asks for the grasp moment alone and reads from the clip how the demonstrator holds the pot: by the handle, with the fingers farther out than its own. Where that handle lies in its own scene it determines from its own cameras, since the demonstration was recorded in another layout. The clip thus supplies what text memory cannot record, and nothing else from the video enters the context.

\Cref{tab:recursion} shows that this pattern holds across all 400 LIBERO-PRO episodes of \cref{tab:main}: the agent opens a few clips per episode, and almost every episode contains a re-entry.

\begin{table}[t]
\centering\small
\setlength{\tabcolsep}{3pt}
\begin{tabular}{@{}lccc@{}}
\toprule
Suite & Episodes & Clip views / episode & With re-entry \\
\midrule
Spatial & 100 & 2.8 & 98\% \\
Object & 100 & 2.3 & 98\% \\
Goal & 100 & 3.9 & 92\% \\
LIBERO-10 & 100 & 10.4 & 98\% \\
\midrule
All & 400 & 4.8 & 97\% \\
\bottomrule
\end{tabular}
\caption{How the agent uses the demonstration in the 400 LIBERO-PRO episodes of \cref{tab:main} (T and S cells pooled). Clip views are re-entries during execution; the last column is the share of episodes with at least one.}
\vspace{-0.2in}
\label{tab:recursion}
\end{table}

\paragraph{Coarse levels first.}
In every episode the agent reads the keyframes ($L_0$) and the phase map ($L_1$/$L_2$) before its first robot action.

\paragraph{Re-entry is frequent and targeted.}
In 97\% of the episodes the agent re-enters the demonstration during execution. It asks for short moments far more often than for whole phases: of the clip views, 40\% are grasp moments, 30\% release moments, 15\% fixture or end-of-task moments and 16\% whole phases (rounded). Re-entries cluster in the second half of an episode (78\%), when the agent is executing contacts rather than planning. Longer tasks call for more of them, from 2.3 per episode on Object to 10.4 on LIBERO-10.

\paragraph{What prompts a re-entry.}
We read the agent's message before each of 100 randomly sampled re-entries. In 68 of them the agent re-enters after a step went wrong, such as a grasp that did not close, a descent that stopped short or an object that slipped. In the other 32 it re-enters to prepare the next contact, most often right before a grasp or a release. The sample covers the 77\% of re-entries that the agent introduces with a message.

\paragraph{Failed episodes.}
These are almost all on LIBERO-10 and contain 21.2 clip views on average against 4.3 in solved ones. In each the agent keeps re-entering at the same kind of step, most often a grasp or a fixture, without resolving it: watching the demonstration again does not change what the robot can do.

\subsection{Robustness on LIBERO-Plus}
\label{sec:plus}

\Cref{tab:liberoplus} reports success on LIBERO-Plus by perturbation dimension. \method{} raises overall success from 86.7\% to 95.8\% and is at least as good as HarnessVLA on every dimension. The largest gain is under changed lighting (76.5 to 94.1), followed by background, robot initial state and camera viewpoint with about 11 points each. Both methods solve every language variant. As on LIBERO-PRO, the gain concentrates in \mbox{LIBERO-10} (29 of 30 variants solved against 22) and Goal (27 against 23; \cref{tab:liberoplus}, bottom). It is below HarnessVLA only on Spatial, by one variant.

\begin{table}[t]
\centering\small
\setlength{\tabcolsep}{1.4pt}
\begin{tabular}{@{}lcccccccc@{}}
\toprule
Dimension & Lay. & Cam. & Rob. & Lang. & Light & Back. & Noise & All \\
\midrule
HarnessVLA & 82.4 & 83.3 & 82.4 & \textbf{100.0} & 76.5 & 88.2 & 94.1 & 86.7 \\
\quad + \method{} & \textbf{88.2} & \textbf{94.4} & \textbf{94.1} & \textbf{100.0} & \textbf{94.1} & \textbf{100.0} & \textbf{100.0} & \textbf{95.8} \\
\midrule
Suite & \multicolumn{2}{c}{Spatial} & \multicolumn{2}{c}{Object} & \multicolumn{2}{c}{Goal} & \multicolumn{2}{c}{L10} \\
\midrule
HarnessVLA & \multicolumn{2}{c}{\textbf{100.0}} & \multicolumn{2}{c}{96.7} & \multicolumn{2}{c}{76.7} & \multicolumn{2}{c}{73.3} \\
\quad + \method{} & \multicolumn{2}{c}{96.7} & \multicolumn{2}{c}{\textbf{100.0}} & \multicolumn{2}{c}{\textbf{90.0}} & \multicolumn{2}{c}{\textbf{96.7}} \\
\bottomrule
\end{tabular}
\caption{Success rate (\%) on LIBERO-Plus by perturbation dimension (top: object layout, camera viewpoint, robot initial state, language, lighting, background and sensor noise) and by suite (bottom). Bold marks the best result in each column.}
\label{tab:liberoplus}
\vspace{-0.2in}
\end{table}

Every perturbation separates the episode from the demonstration, which was recorded in the original scene, camera, light and start pose. The demonstration still helps, which fits how the agent uses it: it takes from a clip how a step is done and localises every object in its own camera images. A new viewpoint or light therefore changes what the agent sees in the episode but not what it learns from the clip. Each dimension has only 17 or 18 variants, so one episode moves a cell by about six points. We therefore read \cref{tab:liberoplus} as evidence that no perturbation breaks \method{}, not as a ranking of the dimensions.

%% file: sections/conclusion.tex
\section{Conclusion}

We presented Recursive Video In-Context Learning (\method{}), which gives an LLM robot agent one demonstration of its task as a hierarchy it navigates: coarse levels ground the plan, and during execution the agent returns to the demonstration at the node of its current sub-goal whenever it needs more detail. The approach needs no training and no change to the policy. The hierarchy is built from the sub-events of the demonstration, so it applies to any task that decomposes into sub-tasks. On LIBERO-PRO it raises the success rate of HarnessVLA from 92.6\% to 96.5\%, most of all on the long LIBERO-10 tasks, where keyframes or the whole video add little. Agents re-enter the demonstration in 97\% of episodes, mostly to see a grasp or a release. On LIBERO-Plus \method{} holds up under all seven kinds of perturbation, and it carries over to a second planner without change. More broadly, treating demonstrations as an environment to be queried, rather than as a prefix, is a natural fit for agents whose information needs change as~they~act.

%% file: sections/limitations.tex
\section*{Limitations}

\method{} assumes one demonstration per base task; when a benchmark rewrites the goal of a task, parts of the demonstration become invalid and must be flagged to the agent. On LIBERO we detect sub-events from the gripper and from fixture joints. Tasks whose sub-tasks leave no such trace in the recorded state, such as pushing, wiping or pouring, need another detector, for example one that reads the end-effector trajectory or asks a vision-language model. When contact is executed by a learned skill whose motion differs from the demonstrator's, the finer levels carry little that the agent can act on.

Each demonstration view stays in context for the rest of the episode, so cost grows with the number of re-entries. All our experiments use LIBERO tasks with a single robot and a single VLA, so other embodiments and demonstration sources remain untested. Our agents are proprietary models accessed through third-party endpoints, whose behaviour may change over time. Our LIBERO-PRO cells hold 50 episodes and each LIBERO-Plus dimension 17 or 18 variants on one seed, so a single episode moves a cell by two to six points. The baselines in \cref{tab:main} come from separate runs, so they cannot be compared with ours episode by episode.

%% file: sections/ethics.tex
\section*{Ethics Statement}
All experiments are conducted in simulation. The LLM agents run shell commands and could read files outside their task; we isolated conditions by restricting directory permissions and audited every shell command afterwards. Deploying such agents on physical robots requires stronger sandboxing and human oversight. Demonstrations are taken from public benchmark datasets.

%% file: sections/appendix.tex
\raggedbottom
\twocolumn[{%
\centering\small
\begin{tabular}{@{}p{0.2\textwidth}p{0.2\textwidth}p{0.52\textwidth}@{}}
\toprule
Tool & Arguments & Returns and intended use \\
\midrule
\var{view\_task\_video} & \var{part} (page, default 1) & $L_0$: eight captioned keyframes per page (third-person and wrist). Called before planning. \\
\var{view\_task\_map} & -- & $L_1$ and $L_2$ as text: duration, phases between sub-events with what each achieved, moments, goal predicates, task objects and regions. Called before planning. \\
\var{view\_demo\_clip} & \var{moment} or \var{phase}; \var{camera} $\in$ \{agentview, wrist, both\} & $L_3$: up to 20 frames of the requested node. Default camera: both for a moment (contact is legible in the wrist view), agentview for a phase (gross motion). Called when about to make a contact, or when a sub-goal stalls. \\
\bottomrule
\end{tabular}
\captionof{table}{The read-only demonstration tools. None of them moves the robot. Tool descriptions instruct the agent not to read positions from the demonstration and to re-localise every object in its own camera images.}
\label{tab:tools}
\vspace{1.2em}

\centering\small
\setlength{\tabcolsep}{4.2pt}
\begin{tabular}{@{}lccccccccc@{}}
\toprule
Planner & Spat-T & Spat-S & Obj-T & Obj-S & Goal-T & Goal-S & L10-T & L10-S & Overall \\
\midrule
\method{} (Codex, GPT-6 Astra, low) & \textbf{100.0} & \textbf{100.0} & \textbf{100.0} & \textbf{100.0} & \textbf{100.0} & \textbf{98.0} & \textbf{90.0} & \textbf{84.0} & \textbf{96.5} \\
\method{} (Claude Code, Opus 5.5) & \textbf{100.0} & \textbf{100.0} & \textbf{100.0} & \textbf{100.0} & 96.0 & 92.0 & \textbf{90.0} & 80.0 & 94.8 \\
\bottomrule
\end{tabular}
\captionof{table}{Success rate (\%) of \method{} on LIBERO-PRO with two planners: GPT-6 Astra at low effort in the Codex CLI and Claude Opus 5.5 in Claude Code. Bold marks the best result in each column.}
\label{tab:planner}
\vspace{1.2em}}]

\section{Demonstration Tools}
\label{app:tools}

\Cref{tab:tools} lists the three tools exposed to the agent. The phase map for the task in \cref{fig:tree} reads, in abridged form:
\begin{quote}\small\ttfamily
phase 1 [0,87) gripper open: approach\\
phase 2 [87,178) gripper closed: carries white\_yellow\_mug\_1 into microwave\_1\_heating\_region\\
phase 3 [178,329) gripper open: pushes the mug in; closes microwave\_1\\
moments: grasp 1 (87), release 1 (178), close microwave\_1 (310)\\
goal: (And (In white\_yellow\_mug\_1 microwave\_1\_heating\_region) (Close microwave\_1))
\end{quote}

\section{Hierarchy Statistics}
\label{app:hier}
Both benchmarks reuse the same 40 demonstrations, one per LIBERO base task. \Cref{tab:hier} summarises their hierarchies. Every demonstration has $K{=}16$ keyframes, rendered at 512\,px and shown as two pages of eight. They include the first and last frames, three frames for every grasp (just before the gripper closes, as it closes, and at the top of the lift) and two for every release (above the target and as the gripper opens). A moment spans 40 frames around a gripper event and 50 around a fixture event, and the last moment covers the final 40 frames of the episode. A clip holds at most 20 frames per camera at 128\,px. In the agent's context a page of keyframes adds about 5k input tokens, the phase map about 2.4k and one clip about 3.1k. We estimate the image levels from the growth of the input between consecutive model calls in the 240 episodes of \cref{tab:recursion}.

\begin{table}[H]
\centering\small
\setlength{\tabcolsep}{4pt}
\begin{tabular}{@{}lcccc@{}}
\toprule
Suite & Length (s) & Phases & Moments & Map tokens \\
\midrule
Spatial & 6.4 & 3.0 & 2.0 & 2.4k \\
Object & 7.8 & 3.2 & 2.2 & 2.3k \\
Goal & 6.8 & 2.7 & 2.0 & 2.2k \\
LIBERO-10 & 14.6 & 4.6 & 3.9 & 2.6k \\
\bottomrule
\end{tabular}
\caption{The demonstration hierarchy per LIBERO suite over its ten base tasks: mean demonstration length, phases and moments, and the median size of the phase map ($L_1$ and $L_2$) in tokens.}
\label{tab:hier}
\end{table}

\section{Swapping the Planner}
\label{app:planner}

\method{} changes nothing that is specific to the planner, so the same demonstration hierarchy and tools serve any agent the harness supports. To check this, we run \method{} with Claude Opus 5.5 in Claude Code on LIBERO-PRO with five seeds per task, as in \cref{tab:main}, and compare it with GPT-6 Astra in \cref{tab:planner}. \method{} with Opus 5.5 reaches 94.8\% overall, 1.7 points below GPT-6 Astra (96.5\%). Both planners solve every episode on Spatial and Object and tie on L10-T, so the gap comes from Goal and L10-S. \method{} with Opus 5.5 still scores above every baseline in \cref{tab:main}. The method thus carries over to a second planner without any change.

\begin{table*}[t]
\centering\small
\begin{tabular}{@{}lcccccc@{}}
\toprule
Method & Time (min) & Model calls & Input tokens (M) & Cached (\%) & Output tokens (k) & Final context (k) \\
\midrule
HarnessVLA & 8.3 & 18 & 1.19 & 88.0 & 3.6 & 91.4 \\
\quad + \method{} & 14.6 & 20 & 1.71 & 88.6 & 4.5 & 109.3 \\
\bottomrule
\end{tabular}
\caption{Median cost of one episode on the 120 LIBERO-Plus variants: wall-clock time, model calls, input tokens and the share served from the cache, output tokens, and the size of the final context.}
\label{tab:cost}
\end{table*}

\section{Cost}
\label{app:cost}
\Cref{tab:cost} compares the cost of an episode with and without the demonstration hierarchy on the 120 LIBERO-Plus variants, which both methods ran in the same time window. \method{} takes 1.8 times the wall-clock time and 1.4 times the input tokens of HarnessVLA. For both methods almost nine in ten input tokens are cache hits, because each turn resends the growing context. Demonstration views make up about 21\% of the final context of a \method{} episode. On LIBERO-PRO a \method{} episode takes a median of 11.0 minutes and 1.7M input tokens, and a LIBERO-10 episode 26.2 minutes and 4.2M.

%% file: custom.bib
@inproceedings{brown2020gpt3,
  title     = {Language Models are Few-Shot Learners},
  author    = {Brown, Tom and Mann, Benjamin and Ryder, Nick and Subbiah, Melanie and others},
  booktitle = {Advances in Neural Information Processing Systems},
  year      = {2020}
}

@inproceedings{min2022rethinking,
  title     = {Rethinking the Role of Demonstrations: What Makes In-Context Learning Work?},
  author    = {Min, Sewon and Lyu, Xinxi and Holtzman, Ari and Artetxe, Mikel and Lewis, Mike and Hajishirzi, Hannaneh and Zettlemoyer, Luke},
  booktitle = {Proceedings of the 2022 Conference on Empirical Methods in Natural Language Processing},
  year      = {2022}
}

@inproceedings{alayrac2022flamingo,
  title     = {Flamingo: a Visual Language Model for Few-Shot Learning},
  author    = {Alayrac, Jean-Baptiste and Donahue, Jeff and Luc, Pauline and Miech, Antoine and others},
  booktitle = {Advances in Neural Information Processing Systems},
  year      = {2022}
}

@inproceedings{liang2023cap,
  title     = {Code as Policies: Language Model Programs for Embodied Control},
  author    = {Liang, Jacky and Huang, Wenlong and Xia, Fei and Xu, Peng and Hausman, Karol and Ichter, Brian and Florence, Pete and Zeng, Andy},
  booktitle = {IEEE International Conference on Robotics and Automation},
  year      = {2023}
}

@article{zhang2026harnessvla,
  title   = {Harness {VLA}: Steering Frozen {VLAs} into Reliable Manipulation Primitives via Memory-Guided Agents},
  author  = {Zhang, Yixian and Zhang, Huanming and Gao, Feng and Li, Xiao and Liu, Zhihao and others},
  journal = {arXiv preprint arXiv:2607.08448},
  year    = {2026}
}

@article{li2026robokharness,
  title   = {{Robo-Harness K1}: Harnessing Robot-Use Agents via Perception Augmentation},
  author  = {Li, Zexi and Zhang, Yehang and Li, Wenqian and Huang, Haojian and Wang, Chenxu and others},
  journal = {arXiv preprint arXiv:2609.29389},
  year    = {2026}
}

@article{liu2026rapid,
  title   = {{RAPID}: Robot Agentic Programming from Demonstrations},
  author  = {Liu, Yuyao and Mao, Jiayuan and Hsu, David and Kaelbling, Leslie Pack and Lozano-P{\'e}rez, Tom{\'a}s},
  journal = {arXiv preprint arXiv:2609.30249},
  year    = {2026}
}

@article{xu2026stellavla,
  title   = {{StellaVLA}: In-Context Structured Demonstration for Generalizable Vision-Language-Action Models},
  author  = {Xu, Siyu and Wang, Yunke and Wang, Zijian and Zhu, Dihao and Xia, Chenghao and others},
  journal = {arXiv preprint arXiv:2608.11671},
  year    = {2026}
}

@article{zhang2026astrarobodojo,
  title   = {An Unexpected Robot Policy: Early Evaluations of {GPT-6 Astra} on {RoboDojo} and Beyond},
  author  = {Zhang, Wenbo and Wang, Kaixuan and Ouyang, Yutao and Huang, Xiaoyu and Li, Liyang and Su, Kailun and others},
  journal = {arXiv preprint arXiv:2609.24170},
  year    = {2026}
}

@article{black2025pi05,
  title   = {$\pi_{0.5}$: a Vision-Language-Action Model with Open-World Generalization},
  author  = {{Physical Intelligence} and Black, Kevin and Brown, Noah and Darpinian, James and Dhabalia, Karan and others},
  journal = {arXiv preprint arXiv:2504.16054},
  year    = {2025}
}

@inproceedings{carion2026sam3,
  title     = {{SAM} 3: Segment Anything with Concepts},
  author    = {Carion, Nicolas and Gustafson, Laura and Hu, Yuan-Ting and Debnath, Shoubhik and Hu, Ronghang and others},
  booktitle = {International Conference on Learning Representations},
  year      = {2026}
}

@inproceedings{liu2023libero,
  title     = {{LIBERO}: Benchmarking Knowledge Transfer for Lifelong Robot Learning},
  author    = {Liu, Bo and Zhu, Yifeng and Gao, Chongkai and Feng, Yihao and Liu, Qiang and Zhu, Yuke and Stone, Peter},
  booktitle = {Advances in Neural Information Processing Systems Datasets and Benchmarks Track},
  year      = {2023}
}

@article{zhou2025liberopro,
  title   = {{LIBERO-PRO}: Towards Robust and Fair Evaluation of Vision-Language-Action Models Beyond Memorization},
  author  = {Zhou, Xueyang and Xu, Yangming and Tie, Guiyao and Chen, Yongchao and Zhang, Guowen and others},
  journal = {arXiv preprint arXiv:2510.03827},
  year    = {2025}
}

@article{kim2024openvla,
  title   = {{OpenVLA}: An Open-Source Vision-Language-Action Model},
  author  = {Kim, Moo Jin and Pertsch, Karl and Karamcheti, Siddharth and Xiao, Ted and Balakrishna, Ashwin and others},
  journal = {arXiv preprint arXiv:2406.09246},
  year    = {2024}
}

@inproceedings{black2025pi0,
  title     = {$\pi_0$: A Vision-Language-Action Flow Model for General Robot Control},
  author    = {Black, Kevin and Brown, Noah and Driess, Danny and Esmail, Adnan and Equi, Michael and others},
  booktitle = {Robotics: Science and Systems},
  year      = {2025}
}

@article{lee2025molmoact,
  title   = {{MolmoAct}: Action Reasoning Models that can Reason in Space},
  author  = {Lee, Jason and Duan, Jiafei and Fang, Haoquan and Deng, Yuquan and Liu, Shuo and others},
  journal = {arXiv preprint arXiv:2508.07917},
  year    = {2025}
}

@article{hung2025nora,
  title   = {{NORA}: A Small Open-Sourced Generalist Vision Language Action Model for Embodied Tasks},
  author  = {Hung, Chia-Yu and Sun, Qi and Hong, Pengfei and Zadeh, Amir and Li, Chuan and others},
  journal = {arXiv preprint arXiv:2504.19854},
  year    = {2025}
}

@inproceedings{zheng2026xvla,
  title     = {{X-VLA}: Soft-Prompted Transformer as Scalable Cross-Embodiment Vision-Language-Action Model},
  author    = {Zheng, Jinliang and Li, Jianxiong and Wang, Zhihao and Liu, Dongxiu and Kang, Xirui and others},
  booktitle = {International Conference on Learning Representations},
  year      = {2026}
}

@article{sun2026atomvla,
  title   = {{AtomVLA}: Scalable Post-Training for Robotic Manipulation via Predictive Latent World Models},
  author  = {Sun, Xiaoquan and Xu, Zetian and Cao, Chen and Liu, Zonghe and Sun, Yihan and others},
  journal = {arXiv preprint arXiv:2603.08519},
  year    = {2026}
}

@article{fu2026capx,
  title   = {{CaP-X}: A Framework for Benchmarking and Improving Coding Agents for Robot Manipulation},
  author  = {Fu, Letian and Yu, Justin and El-Refai, Karim and Kou, Ethan and Xue, Haoru and others},
  journal = {arXiv preprint arXiv:2603.22435},
  year    = {2026}
}

@article{zhang2026rats,
  title   = {Playful Agentic Robot Learning},
  author  = {Zhang, Junyi and Ge, Jiaxin and Yoo, Hanjun and Fu, Letian and Yang, Zihan and others},
  journal = {arXiv preprint arXiv:2606.19419},
  year    = {2026}
}

@article{yu2025rlinf,
  title   = {{RLinf}: Flexible and Efficient Large-Scale Reinforcement Learning via Macro-to-Micro Flow Transformation},
  author  = {Yu, Chao and Wang, Yuanqing and Guo, Zhen and Lin, Hao and Xu, Si and others},
  journal = {arXiv preprint arXiv:2509.15965},
  year    = {2025}
}

@inproceedings{sarthi2024raptor,
  title     = {{RAPTOR}: Recursive Abstractive Processing for Tree-Organized Retrieval},
  author    = {Sarthi, Parth and Abdullah, Salman and Tuli, Aditi and Khanna, Shubh and Goldie, Anna and Manning, Christopher D.},
  booktitle = {International Conference on Learning Representations},
  year      = {2024}
}

@article{liu2024lost,
  title   = {Lost in the Middle: How Language Models Use Long Contexts},
  author  = {Liu, Nelson F. and Lin, Kevin and Hewitt, John and Paranjape, Ashwin and Bevilacqua, Michele and Petroni, Fabio and Liang, Percy},
  journal = {Transactions of the Association for Computational Linguistics},
  volume  = {12},
  year    = {2024}
}

@inproceedings{wang2024videoagent,
  title     = {{VideoAgent}: Long-form Video Understanding with Large Language Model as Agent},
  author    = {Wang, Xiaohan and Zhang, Yuhui and Zohar, Orr and Yeung-Levy, Serena},
  booktitle = {European Conference on Computer Vision},
  year      = {2024}
}

@inproceedings{ahn2022saycan,
  title     = {Do As {I} Can, Not As {I} Say: Grounding Language in Robotic Affordances},
  author    = {Ahn, Michael and Brohan, Anthony and Brown, Noah and Chebotar, Yevgen and Cortes, Omar and others},
  booktitle = {Conference on Robot Learning},
  year      = {2022}
}

@inproceedings{huang2022innermonologue,
  title     = {Inner Monologue: Embodied Reasoning through Planning with Language Models},
  author    = {Huang, Wenlong and Xia, Fei and Xiao, Ted and Chan, Harris and Liang, Jacky and others},
  booktitle = {Conference on Robot Learning},
  year      = {2022}
}

@article{packer2023memgpt,
  title   = {{MemGPT}: Towards {LLMs} as Operating Systems},
  author  = {Packer, Charles and Wooders, Sarah and Lin, Kevin and Fang, Vivian and Patil, Shishir G. and Stoica, Ion and Gonzalez, Joseph E.},
  journal = {arXiv preprint arXiv:2310.08560},
  year    = {2023}
}

@article{wang2023voyager,
  title   = {Voyager: An Open-Ended Embodied Agent with Large Language Models},
  author  = {Wang, Guanzhi and Xie, Yuqi and Jiang, Yunfan and Mandlekar, Ajay and Xiao, Chaowei and Zhu, Yuke and Fan, Linxi and Anandkumar, Anima},
  journal = {arXiv preprint arXiv:2305.16291},
  year    = {2023}
}

@article{wu2021recursively,
  title   = {Recursively Summarizing Books with Human Feedback},
  author  = {Wu, Jeff and Ouyang, Long and Ziegler, Daniel M. and Stiennon, Nisan and Lowe, Ryan and Leike, Jan and Christiano, Paul},
  journal = {arXiv preprint arXiv:2109.10862},
  year    = {2021}
}

@inproceedings{yao2023tot,
  title     = {Tree of Thoughts: Deliberate Problem Solving with Large Language Models},
  author    = {Yao, Shunyu and Yu, Dian and Zhao, Jeffrey and Shafran, Izhak and Griffiths, Thomas L. and Cao, Yuan and Narasimhan, Karthik},
  booktitle = {Advances in Neural Information Processing Systems},
  year      = {2023}
}

@inproceedings{shinn2023reflexion,
  title     = {Reflexion: Language Agents with Verbal Reinforcement Learning},
  author    = {Shinn, Noah and Cassano, Federico and Berman, Edward and Gopinath, Ashwin and Narasimhan, Karthik and Yao, Shunyu},
  booktitle = {Advances in Neural Information Processing Systems},
  year      = {2023}
}

@inproceedings{madaan2023selfrefine,
  title     = {Self-Refine: Iterative Refinement with Self-Feedback},
  author    = {Madaan, Aman and Tandon, Niket and Gupta, Prakhar and Hallinan, Skyler and Gao, Luyu and Wiegreffe, Sarah and others},
  booktitle = {Advances in Neural Information Processing Systems},
  year      = {2023}
}

@inproceedings{wang2025videotree,
  title     = {{VideoTree}: Adaptive Tree-based Video Representation for {LLM} Reasoning on Long Videos},
  author    = {Wang, Ziyang and Yu, Shoubin and Stengel-Eskin, Elias and Yoon, Jaehong and Cheng, Feng and Bertasius, Gedas and Bansal, Mohit},
  booktitle = {Proceedings of the Computer Vision and Pattern Recognition Conference (CVPR)},
  year      = {2025}
}

@article{jain2024vid2robot,
  title   = {{Vid2Robot}: End-to-end Video-conditioned Policy Learning with Cross-Attention Transformers},
  author  = {Jain, Vidhi and Attarian, Maria and Joshi, Nikhil J. and Wahid, Ayzaan and Driess, Danny and Vuong, Quan and others},
  journal = {arXiv preprint arXiv:2403.12943},
  year    = {2024}
}

@inproceedings{yu2018daml,
  title     = {One-Shot Imitation from Observing Humans via Domain-Adaptive Meta-Learning},
  author    = {Yu, Tianhe and Finn, Chelsea and Xie, Annie and Dasari, Sudeep and Zhang, Tianhao and Abbeel, Pieter and Levine, Sergey},
  booktitle = {Robotics: Science and Systems},
  year      = {2018}
}

@article{fu2024icrt,
  title   = {In-Context Imitation Learning via Next-Token Prediction},
  author  = {Fu, Letian and Huang, Huang and Datta, Gaurav and Chen, Lawrence Yunliang and Panitch, William Chung-Ho and Liu, Fangchen and others},
  journal = {arXiv preprint arXiv:2408.15980},
  year    = {2024}
}

@article{dipalo2024kat,
  title   = {Keypoint Action Tokens Enable In-Context Imitation Learning in Robotics},
  author  = {Di Palo, Norman and Johns, Edward},
  journal = {arXiv preprint arXiv:2403.19578},
  year    = {2024}
}

@inproceedings{huang2023voxposer,
  title     = {{VoxPoser}: Composable {3D} Value Maps for Robotic Manipulation with Language Models},
  author    = {Huang, Wenlong and Wang, Chen and Zhang, Ruohan and Li, Yunzhu and Wu, Jiajun and Fei-Fei, Li},
  booktitle = {Conference on Robot Learning},
  year      = {2023}
}

@inproceedings{singh2023progprompt,
  title     = {{ProgPrompt}: Generating Situated Robot Task Plans using Large Language Models},
  author    = {Singh, Ishika and Blukis, Valts and Mousavian, Arsalan and Goyal, Ankit and Xu, Danfei and Tremblay, Jonathan and others},
  booktitle = {IEEE International Conference on Robotics and Automation},
  year      = {2023}
}

@inproceedings{asai2024selfrag,
  title     = {{Self-RAG}: Learning to Retrieve, Generate, and Critique through Self-Reflection},
  author    = {Asai, Akari and Wu, Zeqiu and Wang, Yizhong and Sil, Avirup and Hajishirzi, Hannaneh},
  booktitle = {International Conference on Learning Representations},
  year      = {2024}
}

@inproceedings{jiang2023flare,
  title     = {Active Retrieval Augmented Generation},
  author    = {Jiang, Zhengbao and Xu, Frank F. and Gao, Luyu and Sun, Zhiqing and Liu, Qian and Dwivedi-Yu, Jane and others},
  booktitle = {Proceedings of the 2023 Conference on Empirical Methods in Natural Language Processing},
  year      = {2023}
}

@inproceedings{prasad2024adapt,
  title     = {{ADaPT}: As-Needed Decomposition and Planning with Language Models},
  author    = {Prasad, Archiki and Koller, Alexander and Hartmann, Mareike and Clark, Peter and Sabharwal, Ashish and Bansal, Mohit and Khot, Tushar},
  booktitle = {Findings of the Association for Computational Linguistics: NAACL 2024},
  year      = {2024}
}

@inproceedings{schroeder2025thread,
  title     = {{THREAD}: Thinking Deeper with Recursive Spawning},
  author    = {Schroeder, Philip and Morgan, Nathaniel W. and Luo, Hongyin and Glass, James R.},
  booktitle = {Proceedings of the 2025 Conference of the Nations of the Americas Chapter of the Association for Computational Linguistics: Human Language Technologies (Volume 1: Long Papers)},
  year      = {2025}
}

@article{zhang2025rlm,
  title   = {Recursive Language Models},
  author  = {Zhang, Alex L. and Kraska, Tim and Khattab, Omar},
  journal = {arXiv preprint arXiv:2512.24601},
  year    = {2025}
}

@article{fei2025liberoplus,
  title   = {{LIBERO-Plus}: In-depth Robustness Analysis of Vision-Language-Action Models},
  author  = {Fei, Senyu and Wang, Siyin and Shi, Junhao and Dai, Zihao and Cai, Jikun and Qian, Pengfang and others},
  journal = {arXiv preprint arXiv:2510.13626},
  year    = {2025}
}
